\documentclass[journal]{IEEEtran}
\usepackage{cite}

\ifCLASSINFOpdf
\else
\fi
\usepackage[cmex10]{amsmath}
\usepackage{amssymb} 
\usepackage{color}
\usepackage{algcompatible}
\usepackage[english]{babel}
\usepackage{mathrsfs}
\usepackage{comment}
\usepackage{multicol}
\usepackage{multirow}
\usepackage{booktabs}
\usepackage{algorithm}

\usepackage{graphicx}
\usepackage[colorinlistoftodos]{todonotes}
\usepackage[colorlinks=true, allcolors=blue]{hyperref}

\usepackage{subfigure}

\makeatletter
\newcommand{\removelatexerror}{\let\@latex@error\@gobble}
\makeatother

\begin{document}
%
\title{A Hybrid Approach with Optimization and Metric-based Meta-Learner for Few-Shot Learning}
%
%
%

\author{Duo Wang,~Yu~Cheng,~Mo~Yu,~Xiaoxiao~Guo,
        and~Tao~Zhang,~\IEEEmembership{Senior~Member,~IEEE}
\thanks{Duo Wang and Tao Zhang are with the Department of Automation, Tsinghua University, Beijing 100084, China. (e-mail: d-wang15@mails.tsinghua.edu.cn, taozhang@mail.tsinghua.edu.cn)}
\thanks{Yu Cheng is with Microsoft AI \& Research, Redmond, Washington 98052, USA. (e-mail: yu.cheng@microsoft.com)}
\thanks{Mo Yu and Xiaoxiao Guo are with IBM T.J. Watson Research Center, Yorktown Heights, New York 10562, USA. (e-mail: \{yum, Xiaoxiao.Guo\}@us.ibm.com)}}

%
%

\markboth{Neurocomputing (arXiv extended version)}%
{Shell \MakeLowercase{\textit{et al.}}: Bare Demo of IEEEtran.cls for Journals}
%



\maketitle

\begin{abstract}
Few-shot learning aims to learn classifiers for new classes with only a few training examples per class. Most existing few-shot learning approaches belong to either metric-based meta-learning or optimization-based meta-learning category, both of which have achieved successes in the simplified ``$k$-shot $N$-way'' image classification settings. Specifically, the optimization-based approaches train a meta-learner to predict the parameters of the task-specific classifiers. The task-specific classifiers are required to be homogeneous-structured to ease the parameter prediction, so the meta-learning approaches could only handle few-shot learning problems where the tasks share a uniform number of classes.   
The metric-based approaches learn one task-invariant metric for all the tasks. Even though the metric-learning approaches allow different numbers of classes, they require the tasks all coming from a similar domain such that there exists a uniform metric that could work across tasks. In this work, we propose a hybrid meta-learning model called Meta-Metric-Learner which combines the merits of both optimizations- and metric-based approaches. Our meta-metric-learning approach consists of two components, a task-specific metric-based learner as a base model, and a meta-learner that learns and specifies the base model. Thus our model can handle flexible numbers of classes as well as generate more generalized metrics for classification across tasks. We test our approach in the standard ``$k$-shot $N$-way'' few-shot learning setting following previous works and a new realistic few-shot setting with flexible class numbers in both single-source form and multi-source forms. Experiments show that our approach can obtain superior performance in all settings.
\end{abstract}


%
\IEEEpeerreviewmaketitle

\section{Introduction}
\IEEEPARstart{S}{upervised} deep learning methods have been widely used in visual classification tasks and achieved great success\cite{he2016deep,krizhevsky2012imagenet,NIPS2016_6340,simonyan2014very}. In practice, those methods usually require a large amount of labeled data for model training, in order to make the learned model generalize well.
However, collecting sufficient amount of training data for each task  needs a lot of human work and the process is time-consuming or infeasible for rare classes or classes that
might be hard to observe.

To solve this issue, few-shot learning (FSL)~\cite{li2006one} was proposed, which aims to learn classifiers for new classes with only a few training examples. Generally, two key ideas of few-shot learning are data aggregation and knowledge sharing. First, though each single learning task lacks sufficient annotated data, the union of all the tasks will provide a significantly large amount of labeled data for model training. Therefore the model on a new coming task could benefit from all the previous tasks. Secondly, the experiences of learning model parameters for various of tasks in the past will assist the learning process of the incoming new task. In recent years, deep learning techniques have been successfully exploited for FSL via learning meta-models from a large number of meta-training tasks. Relative methods have been proposed include: 1) learning metric/similarity from multiple few-shot learning tasks with deep networks (metric-based meta-learning) \cite{koch2015siamese,vinyals2016matching}; 
and 2) learning a meta-model on multiple few-shot learning tasks, which could be then used to predict model weights given a new few-shot learning task (optimization-based meta-learning) \cite{kaiser2017learning,ravi2017optimization,munkhdalai2017meta}.

The aforementioned deep few-shot learning models usually are applied to the so-called ''$k$-shot, $N$-way'' scenario, in which each few-shot learning task has the same $N$ number of class labels and each label has $k$ training instances.
However, such ``$N$-way'' simplification is not realistic in real-world few-shot learning applications, because different tasks usually do not have the same number of classes.
Existing optimization-based approaches build on the ''$N$-way'' simplification to let the meta-learner predict weights of homogeneous-structured task-specific networks. If we allow different tasks with different numbers of labels, the task-specific networks will be heterogeneous. Heterogeneous-structured task-specific networks complicate the weight prediction of the meta-learner. To the best of our knowledge, none of the existing optimization-based few-shot learning approaches could resolve this issue.  
Although the metric-based approaches could alleviate the variations on the number of class labels, they suffer from the limitation of model expressiveness: these methods usually learn a task-invariant metric for all the few-shot learning tasks. However, because of the variety of tasks, the optimal metric will also vary across tasks. The learned task-invariant metric may not generalize well if the tasks diverge.

Moreover, in real-world applications, the few-shot learning tasks usually come from different domains or different data sources. For example, for classification of hand-written letter images, we could have images from different languages or alphabets. 
In such few-shot learning scenario, the two aforementioned issues of existing few-shot learning approaches will become more serious: the numbers and meanings of class labels may vary a lot among different tasks, so it will be hard for a meta-learner to learn how to predict weights for heterogeneous neural networks given the few-shot labeled data; and different tasks are not guaranteed to be even closely related to each other, so there will unlikely exist a uniform metric suitable for all the tasks from different domains or data sources.

We propose a hybrid approach which takes the benefits of both metric-based and optimization-based approaches.  The model consists of two main learning modules. The meta learner that operates across tasks uses an optimization-based meta-model to discover good parameters and gradient descent in task-specific base learners. The base learner exploits a metric-based model and parameterizes the task metrics using the weight prediction from the optimization-based meta-learner. Metric-based models are essentially non-parametric models so they are not sensitive to the number of classes. Thus the proposed model is now able to handle unbalanced classes in meta-train and meta-test sets as the usage of metric-based learners as well as to generate task-specific metrics leveraging the weight prediction of the meta-learner given task instances so that the metrics would adapt better to different tasks. Our model exploits metric-learning based model to perform classification but learns it via training an optimization-based meta-model, so we call it Meta-Metric-Learner. In this work, we propose two types of Meta-Metric-Learner, i.e. Meta Matching Network (MMN) \cite{vinyals2016matching} and Meta Prototypical Network (MPN) \cite{}, which exploit the same model as the meta learner but two different metric-based models (Matching Network and Prototypical Network respectively) as the base learner.

We make the following contributions: 1) we design the meta-metric-learning method that is able to learn task-specific metrics via training a meta-learner; 2) we propose training methods towards the meta-metric-learner in both single-source and multi-source settings; 3) we evaluate our model on several benchmark datasets with various baselines.
These contributions make our approach suit better to the few-shot learning problem. We test our approach in both classic ``$k$-shot $N$-way''  few-shot learning setting following previous work
and a new but more realistic few-shot setting with flexible label numbers using single-source and multi-source training data. 
Experiments show that our approach attains superior performance on all of the settings.

This work is an extension of \cite{cheng2017few} in several ways. Firstly, we exploit different kinds of models in both of the two component learning modules in our approach. In specific, we use Meta-SGD  as meta-learner to predict the parameters of metric-based learners. While, in \cite{cheng2017few}, LSTM-based model is used. And both Matching Networks and Prototypical Networks are selected as metric-based learner in our experiment. Secondly, we conduct all of the experiments with image datasets other than text ones and the experiment settings are different. Besides, a more detailed presentation of our approach and discussion of the most recent related works are given in this paper.

The rest of the paper is organized as follows. In Section 2, related works and literature associated with few-shot learning are reviewed. Section 3 describes the proposed model in detail. A series of experiments are conducted in Section 4. Finally, the conclusion and future work are discussed in Section 5.

\section{Background}
Few-shot learning~\cite{li2006one, miller2000learning} aims to learn classifiers for new classes with only a few training examples per class. Bayesian Program Induction~\cite{lake2015human}, which can be seen as the most pioneering work in recent years, represents concepts as simple programs that best explain observed examples under a Bayesian criterion and reaches human level error on few-shot learning alphabet recognition tasks. This work is a successful instance of meta-learning, which has a long history~\cite{thrun1998learning,schmidhuber1997shifting}. 
The key idea of meta-learning is framing the learning problem at two levels: the lower level is the quick acquisition of knowledge from each separate task presented and the higher level is accumulating knowledge to learn the similarities and differences across all tasks. 
Since then, great progress has been made in few-shot learning. Most existing few-shot learning approaches belong to either metric-based meta-learning or optimization-based meta-learning category. For metric-based works, \cite{SantoroBBWL16} exploits Neural Turing Machine (NTM)~\cite{GravesWD14}, a famous memory-augmented neural network, to few-shot learning problem and introduces a new attention-based memory accessing method to rapidly assimilate new data used for accurate predictions about new classes. Siamese neural networks rank similarity between inputs~\cite{koch2015siamese}. Matching Networks~\cite{vinyals2016matching} introduce a trainable k-nearest neighbors algorithm to map a small labeled supporting set and an unlabeled example to its label, obviating the need for fine-tuning to adapt to new class types. Prototypical Networks~\cite{NIPS2017_6996} perform classification by computing Euclidean distances to prototype representations of each class. Generative Adversarial Residual Pairwise Networks~\cite{MehrotraD17} exploit deep residual modules in the pairwise network and regularize it with an adversarial training strategy. ~\cite{yang2018rn} 
For optimization-based works, a recent approach~\cite{andrychowicz2016learning} casts the hand-designed optimization algorithm as a learning problem, and trains an LSTM-based meta-learner to predict model parameters. The LSTM-based meta-learner is then applied to few-shot learning tasks~\cite{ravi2017optimization} by training over a bunch of hand-designed few-shot learning tasks.
Model-Agnostic Meta-Learning(MAML)~\cite{MAMLFinnAL17} explicitly trains the initial parameters of CNN model such that a small number of gradient update steps with a small amount of training data from a new task will produce good generalization performance on that task. Meta-SGD~\cite{MetaSGDLiZCL17} extends the idea from \cite{MAMLFinnAL17} by learning to learn not just the learner initialization, but also the learner update direction and learning rate. TAML~\cite{jamal2018taml} presents an entropy-based approach to avoid a biased meta-learner and improve its generalizability to new tasks. \cite{finn2018probabilisticMAML} and \cite{kim2018bayesianMAML} extend MAML to probabilistic forms.

Besides, Deep Meta-Learning~\cite{Zhou18l2lconcept} proposes to perform few-shot learning on high-level representation space rather than instance space. \cite{garcia2018fewshot} considers few-shot learning as supervised message passing task and generalizes several proposed models with graph-based neural networks. Some researchers try to alleviate the scarcity of training data by data augmentation. \cite{Chen18} proposes to augment instance semantic features using a novel auto-encoder network dual TriNet. \cite{oreshkin2018tadam} proposes to scale the distance metric with alearnable parameter. They also define a dynamic feature extractor with parameters predicted from a task representation and a task embedding network. \cite{rusu2018metalearning} tries to learn a data-dependent latent generative representation of model parameters, and performing gradient-based meta-learning in this low-dimensional latent space to tackle the high-dimensional problem in optimization-based meta-learning methods.

Our idea is similar to that of \cite{NIPS2018Tyler}. Both of the works propose to generate task-specific metric that is more adapted to new few-shot learning tasks. However, \cite{NIPS2018Tyler} uses basic fine-tune to adjust the metric-based models, while our paper exploits a meta-learner, Meta-SGD, to calculate the parameters of the metric-learner. \cite{NIPS2018Tyler} is simple and easy to implement since they don’t use additional models to adapt the metric. Our method is more flexible and general. Meta-SGD can be seen as a trainable fine-tune, as the initial parameters and the learning rate are meta-learned, not set by hand. We may exploit other more sophisticated optimization-based meta-learner such as probabilistic MAML \cite{finn2018probabilisticMAML} \cite{kim2018bayesianMAML}, and LEO\cite{rusu2018metalearning}, which is one of our future extensions of this work. In this meaning, the work of few-shot learning in \cite{NIPS2018Tyler} can be seen as a special case of our paper.

Among the few-shot learning methods, Matching Network, Prototypical Networks, MAML, and Meta-SGD are closely related to our method. The remaining of this Section will provide more details on these  few-shot learning approaches. 

\subsection{Few-shot Learning Problem Definition}
In the typical machine learning problem setting, a classification task $\mathcal{T}$ contains a supporting dataset(or training dataset) $S$ to optimize model parameters and a test dataset $T$ to evaluate model performance. For a $k$-shot, $N$-class classification task, the supporting dataset consists of $k$ labeled samples for each of $N$ classes, i.e. there are total $k*N$ samples in the supporting dataset, and the test dataset contains a number of samples of the same $N$ classes for evaluation. In the few-shot learning setting, $k$ is a very small value(we consider k less than 5 in this work), meaning each supporting dataset in classification tasks will contain few labeled examples.

Recently-proposed methods formulates few-shot learning as a meta-learning problem. Under this thought, we have different meta-sets for meta-training, meta-validation, and meta-testing($\mathscr{D}_{meta-train}$, $\mathscr{D}_{meta-validate}$ and $\mathscr{D}_{meta-test}$, respectively), each of which contains a certain number of few-shot learning tasks described above, all drawn from task distribution $p(\mathcal{T})$. With $\mathscr{D}_{meta-train}$, we train a meta-learner to generate good task-specific metrics for few-shot learning tasks and evaluate its generalization performance on $\mathscr{D}_{meta-test}$. $\mathscr{D}_{meta-validate}$ is used to select good hyper-parameters. Note that in this work, tasks in different meta-sets may contain different numbers of classes following real few-shot learning scenarios. 

\subsection{Matching Network(MN) and Prototypical Network(PN)}
\textbf{Matching Networks}~\cite{vinyals2016matching} consist of a neural network as embedding function and an augmented memory. The embedding function, $f()$, maps an input $x \in X$ to a $d$-length vector, i.e., $f: X \rightarrow \mathbb{R}^{d}$. The augmented memory stores a supporting set $S=\{(x_{i},y_{i})\}^{|S|}_{i=1}$, where $x_{i}$ is supporting instance and $y_{i}$ is its corresponding one-hot label. The Matching Networks explicitly define a classifier conditioned on the supporting set. For any new data $\hat{x}$, the Matching Networks predict its label via a similarity function $\alpha(.,.)$ between the instance $\hat{x}$ and  the supporting set $S$:
\begin{equation}
\label{equ1}
y = P(.|\hat{x}, S) = \sum_{i=1}^{|S|} \alpha(\hat{x}, x_{i};\theta) y_{i}.
\end{equation}
Specifically, we define the similarity function to be a softmax distribution given some kind of distance between the testing instance $\hat{x}$ and the supporting instance $x_{i}$, {\emph i.e.}, $\alpha(\hat{x}, x_{i};\theta) = \exp[d(f(\hat{x}),  f(x_{i}))] / \sum_{j=1}^{k} \exp[d(f(\hat{x}), f(x_{j}))]$, where $\theta$ are the parameters of the embedding function $f$ and $d:\mathbb{R}^{d}\times\mathbb{R}^{d} \rightarrow [0,+\infty)$ is distance function. Thus, $y$ is a valid distribution over the supporting set's labels $\{y_{i}\}_{i=1}^{|S|}$. 
Here $f$ is parameterized as deep convolutional neural networks for image tasks and cosine distance is adopted as the distance function.

For the training of Matching Networks, we first sample a few-shot learning task $\mathcal{T}$ with a supporting set $S$ and a test set $T$ from $\mathscr{D}_{meta-train}$.  
The objective function to optimize the embedding parameters is to minimize the prediction error of the testing samples given the supporting set as follows: 
\begin{equation}
\mathop{\mathbb{E}}_{\mathcal{T}\sim p(\mathcal{T})} \Big[ \mathop{\mathbb{E}}_{S,T\sim \mathcal{T}} \big[ \sum_{(x,y)\in T} \log(P(y|x,S;\theta))\big] \Big].
\end{equation}
The parameters of the embedding function, $\theta$, are optimized via stochastic gradient descent methods. 

\textbf{Prototypical Networks}~\cite{NIPS2017_6996} can be seen as a variation of Matching Networks, which perform classification in a different way from Eq.(\ref{equ1}):
\begin{equation}
\label{equ2}
y = P(.|\hat{x}, S) = \sum_{i=1}^{N} \alpha(\hat{x}, c_{i};\theta) y_{i}.
\end{equation}
Here $c_{i}$ is the mean embedded vector of the supporting samples belonging to class $i$: 
\begin{equation}
c_{i} = \frac{1}{|S_{i}|}\sum_{x_{j}\in S_{i}}f(x_{j})
\end{equation}
$S_{i}$ denotes the set of examples labeled with class $i$ in the given supporting set. $c_{i}$ is called prototype and can be considered as a representation of its belonging class. We choose Euclidean distance as the distance function in $\alpha(\hat{x}, c_{i};\theta)$ as it works better for image few-shot learning tasks\cite{NIPS2017_6996}.

\subsection{MAML and Meta-SGD}
\textbf{MAML}~\cite{MAMLFinnAL17} does not use an explicit learnable model to perform update of learner's weights like \cite{andrychowicz2016learning} and \cite{ravi2017optimization}. It is just simply based on the gradient-descent method.
The underlying key idea is to train the learner’s initial parameters such that the learner has maximal performance on a new few-shot learning task after the parameters have been updated through one or more gradient steps computed with a small amount of supporting samples from that new task. Assume that the learner can be represented by a parametrized function $f_{\theta}$ with initial parameters $\theta$. Given a new task $\mathcal{T}_{i} \sim p(\mathcal{T})$ with a supporting set $S_{i}$ and a testing set $T_{i}$, the initial parameters $\theta$ are updated to $\theta^{'}$ using one or more gradient descent steps calculated by $S_{i}$ to adapt to the new task. Take one update step as an example:
\begin{equation}
\theta^{'} = \theta - \alpha \nabla_{\theta}\mathcal{L}_{S_{i}}(\theta)
\end{equation}
The initial parameters are trained so that the learners with updated parameters $f_{\theta^{'}}$ will have maximal performance across several new tasks sampled from $p(\mathcal{T})$. The meta-training objective is as follows:
\begin{equation}
\min_{\theta} E_{\mathcal{T} \sim p(\mathcal{T})}[\mathcal{L}_{T}(\theta^{'})]= E_{\mathcal{T} \sim p(\mathcal{T})}[\mathcal{L}_{T}(\theta-\alpha \nabla_{\theta}\mathcal{L}_{S}(\theta))]
\end{equation}
Training is implemented using SGD as follows:
\begin{equation}
\theta \leftarrow \theta - \beta \nabla_{\theta}E_{\mathcal{T} \sim p(\mathcal{T})}[\mathcal{L}_{T}(\theta^{'})]
\end{equation}

\textbf{Meta-SGD}~\cite{MetaSGDLiZCL17} extends the idea of MAML for a little bit. They vectorize the step size $\alpha$ in MAML with equal dimension to learner's parameters and make it trainable as well. So Meta-SGD learns not only the initial parameters but also the update direction and the update rate. The meta-training objective is:
\begin{equation}
\min_{\theta, \alpha} E_{\mathcal{T} \sim p(\mathcal{T})}[\mathcal{L}_{T}(\theta^{'})]= E_{\mathcal{T} \sim p(\mathcal{T})}[\mathcal{L}_{T}(\theta-\alpha \circ \nabla_{\theta}\mathcal{L}_{S}(\theta))]
\end{equation}

In the above content, we only consider one gradient step, but it is a straightforward extension to use multiple steps in experiments.

\begin{figure*}
	\begin{center}
		\makebox[\textwidth]{\includegraphics[width=0.85\paperwidth]{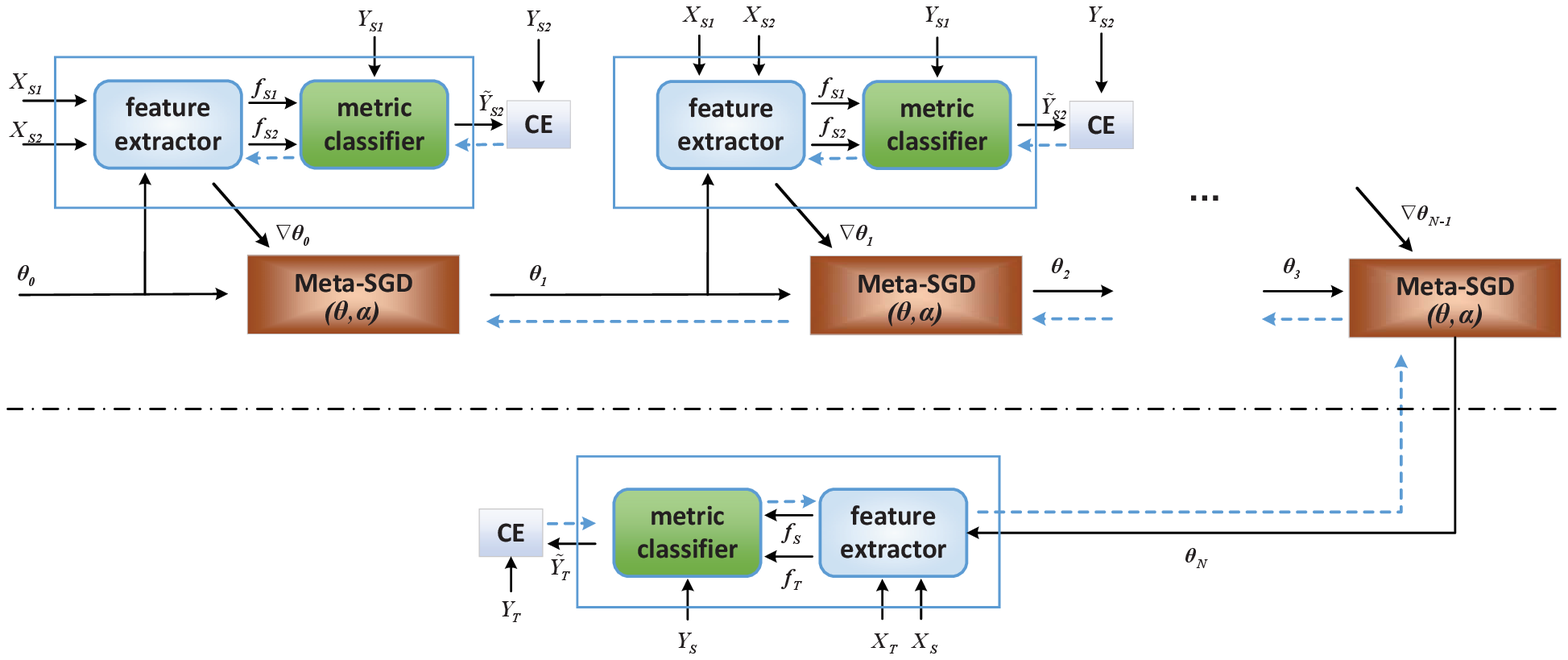}}
	\end{center}
	\caption{Illustration of the training procedure for our model. ($X_{S}$, $Y_{S}$) are samples of the supporting set(or training set) $S$ and ($X_{T}$, $Y_{T}$) are samples of the testing set $T$ in a few-shot learning task. ($X_{S1}$, $Y_{S1}$) and ($X_{S2}$, $Y_{S2}$) are samples of two subsets of $S$ respectively used for the forward pass. The blue rounded rectangles indicate feature extractors parameterized as deep convolutional neural networks(CNN) with parameters $\theta$ and $f$ is the output feature. The green rounded rectangles are metric-based classifiers which can be k-nearest neighbor classifiers or prototype-based classifiers in our model. $\tilde{Y}$  is the prediction of the classifier. CE means cross-entropy loss. The red rectangles are Meta-SGD modules which take as input the gradient to update the model parameters. The black arrows indicate forward pass and the blue dash arrows indicate backward pass. We use ($X_{S1}$, $Y_{S1}$), ($X_{S2}$, $Y_{S2}$) and the Meta-SGD module to update the parameters of the feature extractor for $N$ steps and then evaluate on ($X_{T}$, $Y_{T}$) of $T$. Meta-SGD module is trained to minimize the cross-entropy loss between $\tilde{Y}_{T}$ and ${Y}_{T}$ (should be viewed in color). } 
	\label{fig:arc}
\end{figure*}

\section{Meta-Metric-Learner for Few-Shot Learning}
In this section, we provide the details of our Meta-Metric-Learner and its training objective. We first describe the Meta-Metric-Learner in a single-source setting. After that, we show it is easy to generalize the model in a multi-source learning setting, which relates to retrieving auxiliary sets from other sources/tasks.

\paragraph{Meta-Metric-Learner}
Following the setting in previous few-shot learning works, we construct three meta-datasets, i.e. $\mathscr{D}_{meta-train}$, $\mathscr{D}_{meta-validate}$ and $\mathscr{D}_{meta-test}$. Each of the meta-dataset consists of a number of few-shot learning tasks. In these previous works, the few-shot learning tasks in the three meta-dataset all have the same number of classes. In our experiment setting, the number of classes in $\mathscr{D}_{meta-test}$ is the same as that in $\mathscr{D}_{meta-validate}$ but could be different from that in $\mathscr{D}_{meta-train}$, taking real scenarios into consideration. Although the CNN base learner used in \cite{ravi2017optimization}, \cite{MAMLFinnAL17} and \cite{MetaSGDLiZCL17} is powerful to model image data, it lacks the ability to handle unbalanced classes in train and test datasets in a straightforward way. On the other hand, metric-learning-based models, as trainable non-parametric algorithms by nature, can generalize easily to new datasets, which contain samples from different numbers of classes. Hence, we apply the Meta-SGD in \cite{MetaSGDLiZCL17} for few-shot learning tasks, but replace the CNN with metric-learning based classification model as the base learner, so that it can tackle class-unbalanced few-shot learning problems. Here, we propose two types of Meta-Metric-Learner, i.e. Meta Matching Network (MMN) and Meta Prototypical Network (MPN), which exploit two different kinds of metric-based models(Matching Network and Prototypical Network respectively) as the base learner.
Suppose we have a Meta-Metric-Learner containing a metric-learner $M(.;\theta)$ with parameters $\theta$ as the base learner and the initial parameters are $\theta_{0}$. The Meta-SGD model can be denoted as $R(.;\theta_{0}, \alpha)$, where $\alpha$ is update step size. 

\paragraph{Meta-Metric-Learner of Single-Source Form} 
We use the meta-training set $\mathscr{D}_{meta-train}$ to train our Meta-Metric-Learner. Specifically, we first sample a few-shot learning task $\mathcal{T}_{i}$ from $\mathscr{D}_{meta-train}$, which contains a supporting dataset $S_{i}$ and a testing dataset $T_{i}$, with all sample labels known. At gradient step $n$, the base metric-learner with parameters $\theta_{n}^{i}$ takes $S_{i}$ as input to calculate classification loss and its gradient w.r.t $\theta_{n}^{i}$. However, the original optimization-based meta-learning approach cannot be applied directly in our model due to the fact that our base learner is metric-based. Metric-learner predicts labels of query samples by exploiting their similarity with labeled supporting samples, i.e. metric-learner itself need two separate datasets for the learning procedure. To tackle this problem, we propose to divide $S_{i}$ into two subsets, denoted as $S_{i1}$ and $S_{i2}$, and use them as query set and supporting set respectively for the metric-learner. Thus, the classification loss can be expressed as:
\begin{equation}
\label{equ9}
\mathcal{L}_{n}^{i} = \mathcal{L}(M(X_{Si1}, X_{Si2}, Y_{Si1};\theta_{n}^{i}), Y_{Si2})
\end{equation}

($X_{Si1}$, $Y_{Si1}$) and ($X_{Si2}$, $Y_{Si2}$) are samples of two subsets $S_{i1}$ and $S_{i2}$ respectively.
Then the Meta-SGD updates metric learner's parameters using the basic gradient-descent method:
\begin{equation}
\theta_{n+1}^{i} = \theta_{n}^{i} - \alpha \circ \nabla_{\theta}\mathcal{L}_{n}^{i}(\theta_{n}^{i})
\end{equation}
After this procedure is repeated for $N$ steps, Meta-SGD updates the base metric learner parameters to $\theta_{N}^{i}$. We make predictions about samples in $T_{i}$ with the updated metric learner and supporting set $S_{i}$ and get evaluation loss:
\begin{equation}
\mathcal{L}_{N}^{i} = \mathcal{L}(M(X_{Si}, X_{Ti}, Y_{Si};\theta_{N}^{i}), Y_{Ti})
\end{equation}
The evaluation loss across all few-shot learning tasks from $\mathscr{D}_{meta-train}$ is minimized to optimize the parameters of the Meta-SGD:
\begin{equation}
\theta_{0}^{*}, \alpha^{*} = \arg\min_{\theta_{0}, \alpha} E_{\mathcal{T}_{i}}[\mathcal{L}_{N}^{i}]
\end{equation}

The overall architecture of our Meta-Metric-Learner is shown in Figure \ref{fig:arc} and the meta-training procedure is given in Algorithm \ref{alg1}.

Note that each of the two subsets, $S_{i1}$ and $S_{i2}$, needs to contain different samples of every class in the few-shot learning task. This means for each class, we need more than one labeled data from $S_{i}$, i.e. the method can only directly handle k-shot problems with $k\geq2$.

\paragraph{Multi-Source Form with Auxiliary Data}
In order to directly handle one-shot learning problems in our model, we propose to borrow instances from other data sources to augment the original meta-training dataset. This makes our method essentially extend to a multi-source learning process. Specifically, we construct an auxiliary meta-set $\mathscr{D}_{aux}$ which contains a number of learning tasks and use it to calculate the classification loss and update the parameters of the base metric learner. Given a one-shot learning task $\mathcal{T}_{i}$ sampled from $\mathscr{D}_{meta-train}$, we randomly choose an auxiliary learning task $\mathcal{T}_{aux-i}$ from $\mathscr{D}_{aux}$. $\mathcal{T}_{aux-i}$ consists of a supporting set $S_{aux-i}$ and a testing set $T_{aux-i}$. The classification loss now is different from Eq.\ref{equ9}:
\begin{equation}
\mathcal{L}_{n}^{i} = \mathcal{L}(M(X_{Saux-i}, X_{Taux-i}, Y_{Saux-i};\theta_{n}^{i}), Y_{Taux-i})
\end{equation}
The update of the metric-learner's parameters and the training of the meta-model is similar to single-source setting described above. Algorithm \ref{alg2} shows the detail of multi-source meta-metric-learning. In fact, this idea is partly motivated by transfer learning, where data in a source domain are used to acquire knowledge to facilitate learning in a related target domain. Here we apply the thought of transfer learning in meta-learning setting. The meta-learner is trained to extract cross-source knowledge from the auxiliary meta-set that is transferable to the target meta-set.

Next, we discuss how to construct the auxiliary set $\mathscr{D}_{aux}$. The intuition is, in many real-world applications, we can get few-shot learning data from multiple data sources, such as images of hand-written symbols from different alphabets. Such data from multiple sources could increase the training data for our few-shot learning method.  
However, there is rarely a guarantee that the above data sources are related to each other. 
When the auxiliary data are from an unrelated source, it will be difficult for the few-shot learning methods to learn a good metric or a good meta-learner. In this case, when adding more significantly unrelated auxiliary data, the performance may decrease. To overcome this difficulty, given a target data source for few-shot learning, we use the following approach to select related data sources similarly with \cite{Yu2018Diverse}.

Consider a list of $n$ data sources (such as a list of alphabets in hand-writing recognition) $\{\mathcal{S}^1, \mathcal{S}^2, ..., \mathcal{S}^n\}$. From each data source $\mathcal{S}^i$ we can sample a meta-dataset $\mathscr{D}_{i}$ containing a number of few-shot learning tasks. Because datasets in few-shot learning tasks are too small to reflect any statistical relatedness among them, our approach deal with the problem at the task-level with the following steps: (1) For each data resource $\mathcal{S}^i$ we use the sampled meta-dataset $\mathscr{D}_{i}$ to train a metric learner $M^i$ on it. (2) For the target data source $\mathcal{S}^{target}$, we also sample a group of tasks and apply each model $M^i$ to get the classification accuracy $acc_{i\rightarrow target}$. Note that the accuracy scores are usually low but their relative magnitudes can reflect the relatedness between different sources to $\mathcal{S}^{target}$. (3) Finally we select the top $s$ sources $\mathcal{S}^i$ with the highest scores $acc_{i\rightarrow target}$ to construct the auxiliary set $\mathscr{D}_{aux}$.

\begin{figure}[!t]
\removelatexerror
\begin{algorithm}[H]
\caption{Meta-Metric-Learner Meta-Training in Single-Source Setting}
\label{alg1}
\begin{algorithmic}[1]
\REQUIRE ~~\\ Meta-train set $\mathscr{D}_{meta-train}$, 
Metric learner $M$ with parameters $\theta$, Meta-Learner Meta-SGD $R$ with parameters $(\theta_{0}, \alpha)$ 
\STATE $\theta_{0}$ $\leftarrow$ random initialization
\WHILE{not done}
\FOR{all $\mathcal{T}_i$ in $\mathscr{D}_{meta-train}$}
\STATE supporting dataset $S_i$,  testing dataset $T_i$ $\leftarrow$ task $\mathcal{T}_i$
\STATE $S_{i1}$, $S_{i2}$ $\leftarrow$ equally split $S_{i}$ into two subsets
\FOR{$n=0,N-1$}
\STATE $X_{Si1}$, $Y_{Si1}$ $\leftarrow$ sampled from $S_{i1}$
\STATE $X_{Si2}$, $Y_{Si2}$ $\leftarrow$ sampled from $S_{i2}$
\STATE $\mathcal{L}_{n}^{i} \leftarrow \mathcal{L}(M(X_{Si1}, X_{Si2}, Y_{Si1};\theta_{n}^{i}), Y_{Si2})$
\STATE $\theta_{n+1}^{i} \leftarrow \theta_{n}^{i} - \alpha \circ \nabla_{\theta}\mathcal{L}_{n}^{i}(\theta_{n}^{i})$
\ENDFOR
\STATE $X_{Si}$, $Y_{Si}$ $\leftarrow$ all samples from $S_{i}$
\STATE $X_{Ti}$, $Y_{Ti}$ $\leftarrow$ all samples from $T_{i}$
\STATE $\mathcal{L}_{N}^{i} \leftarrow \mathcal{L}(M(X_{Si}, X_{Ti}, Y_{Si};\theta_{N}^{i}), Y_{Ti})$

\ENDFOR
\STATE Updating $\theta_{0}$ and $\alpha$ to minimize $E_{\mathcal{T}_{i}}[\mathcal{L}_{N}^{i}]$
\ENDWHILE
\end{algorithmic}
\end{algorithm}
\end{figure}

\begin{figure}[!t]
\removelatexerror
\begin{algorithm}[H]
\caption{Meta-Metric-Learner Meta-Training in Multi-Source Setting}
\label{alg2}
\begin{algorithmic}[1]
\REQUIRE ~~\\ Meta-train set $\mathscr{D}_{meta-train}$, auxiliary dataset $\mathscr{D}_{aux}$,   
Metric learner $M$ with parameters $\theta$, Meta-Learner Meta-SGD $R$ with parameters $(\theta_{0}, \alpha)$ 
\STATE $\theta_{0}$ $\leftarrow$ random initialization
\WHILE{not done}
\FOR{all $\mathcal{T}_i$ in $\mathscr{D}_{meta-train}$}
\STATE supporting dataset $S_i$,  testing dataset $T_i$ $\leftarrow$ task $\mathcal{T}_i$
\STATE $S_{aux-i}$, $T_{aux-i}$ $\leftarrow$ task $\mathcal{T}_{aux-i}$ sampled from $\mathscr{D}_{aux}$ 
\FOR{$n=0,N-1$}
\STATE $X_{Saux-i}$, $Y_{Saux-i}$ $\leftarrow$ sampled from $S_{aux-i}$
\STATE $X_{Taux-i}$, $Y_{Taux-i}$ $\leftarrow$ sampled from $T_{aux-i}$
\STATE $\mathcal{L}_{n}^{i} \leftarrow \mathcal{L}(M(X_{Saux-i}, X_{Taux-i}, Y_{Saux-i};\theta_{n}^{i}), Y_{Taux-i})$
\STATE $\theta_{n+1}^{i} \leftarrow \theta_{n}^{i} - \alpha \circ \nabla_{\theta}\mathcal{L}_{n}^{i}(\theta_{n}^{i})$
\ENDFOR
\STATE $X_{Si}$, $Y_{Si}$ $\leftarrow$ all samples from $S_{i}$
\STATE $X_{Ti}$, $Y_{Ti}$ $\leftarrow$ all samples from $T_{i}$
\STATE $\mathcal{L}_{N}^{i} \leftarrow \mathcal{L}(M(X_{Si}, X_{Ti}, Y_{Si};\theta_{N}^{i}), Y_{Ti})$

\ENDFOR
\STATE Updating $\theta_{0}$ and $\alpha$ to minimize $E_{\mathcal{T}_{i}}[\mathcal{L}_{N}^{i}]$
\ENDWHILE
\end{algorithmic}
\end{algorithm}
\end{figure}

\section{Experiments and Results}
In this section, we conduct experiments with $k$-shot learning in both single-source and multi-source settings. The experiments are conducted on five popular image datasets, comparing Meta-Metric-Learner against several baselines. We first describe the datasets, experimental settings, and baseline models. 

\paragraph{Datasets} The five datasets are MiniImagenet, Caltech-256, Cifar-100, Cub-200, and Omniglot. We use the first four datasets in the single-source setting and Omniglot in the multi-source setting.

1) \textbf{MiniImagenet}: The MiniImagenet dataset, first used in \cite{vinyals2016matching}, consists of 60,000 color images of 100 classes, with 600 images per class. For our experiments, we use the same splits as \cite{ravi2017optimization} to enable the comparison
with previous methods. Their splits use a different set of 100 classes, which are divided into three disjoint subsets: 64 classes for meta-training, 16 classes for meta-validation, and 20 classes for meta-testing.

2) \textbf{Caltech-256}: The Caltech-256 dataset \cite{griffin2007caltech} is a successor to the well-known dataset Caltech-101. It contains totally 30,607 color images of 256 classes. We split it into three subsets: 150, 56, and 50 classes for meta-training, meta-validation, and meta-testing, respectively as \cite{Zhou18l2lconcept}. 

3) \textbf{Cifar-100}: The CIFAR-100 dataset \cite{krizhevsky2009learning} contains 60,000 color images of 100 fine-grained categories, and 20 coarse-level categories, which are both in size of 32x32. We use 64, 16, and 20 categories classes for meta-training, meta-validation, and meta-testing, respectively. 

4) \textbf{Cub-200}: The CUB-200 dataset \cite{wah2011caltech} contains 11,788 color images of 200 different bird species. We use 140 classes for meta-training, 20 classes for meta-validation, and test on the remaining 40 classes. In this fine-grained dataset, images' differences between very similar classes are usually so subtle that they can hardly be recognized even by humans.

5) \textbf{Omniglot}: The data comes with a standard split of 30 training alphabets with 964 classes and 20 evaluation alphabets with 659 classes. Each of these was hand drawn by 20 different people. Each data source corresponds to an alphabet here.

For Cifar-100, we use the images of the original size, i.e. 32x32. For Omniglot, we resize the images to the size of 28x28. For the other three datasets, we resize the images to 84x84.

\begin{table*}
	\footnotesize
	\caption{Average classification accuracies on miniImageNet with different approaches in single-source form.}
	\label{tab:nlu:single1}
	\centering
	\makebox[\textwidth]{
		\begin{tabular}{cccc|cccc}
			\toprule
			\multirow{2}{*}{\textbf{Model}} & \multirow{2}{*}{\textbf{Model Type}} & \multicolumn{2}{c|}{\textbf{5 class}} & \multicolumn{2}{c}{\textbf{5 vs. 3}} & \multicolumn{2}{c}{\textbf{3 vs. 5}}\\
			\cline{3-8}
			& & 2-shot & 4-shot & 2-shot & 4-shot & 2-shot & 4-shot\\
			\midrule
			Meta-SGD & - & 49.89$\pm$0.73 & 56.28$\pm$0.68 & - & - & - & - \\
			\midrule
			Matching Network & Basic (No FCE) & 53.16$\pm$0.69 & 59.66$\pm$0.69 & 66.78$\pm$1.07 & 72.53$\pm$0.84 & 50.27$\pm$0.69 & 57.76$\pm$0.66\\
			Matching Network fine-tune & Basic (No FCE) & 53.92$\pm$0.75 & 60.82$\pm$0.66 & \textbf{69.21$\pm$0.91} & \textbf{74.1$\pm$0.84} & 50.71$\pm$0.73 & 58.45$\pm$0.65\\
			\textbf{Meta Matching Network} & Basic (No FCE) & \textbf{53.92$\pm$0.68} & \textbf{60.94$\pm$0.67} & 67.14$\pm$0.97 & 73.91$\pm$0.83 & \textbf{50.99$\pm$0.68} & \textbf{59.31$\pm$0.65}\\
			\midrule
			Prototypical Network & Euclid. & 50.89$\pm$0.75 & 57.87$\pm$0.70 & 64.56$\pm$0.97 & 71.76$\pm$0.89 & 48.02$\pm$0.76 & 55.57$\pm$0.72\\
			Prototypical Network fine-tune & Euclid. & \textbf{52.47$\pm$0.71} & \textbf{60.68$\pm$0.83} & \textbf{66.48$\pm$0.98} & \textbf{72.52$\pm$0.85} & \textbf{49.51$\pm$0.71} & \textbf{57.64$\pm$0.70}\\
			\textbf{Meta Prototypical Network} & Euclid. & 51.95$\pm$0.68 & 58.44$\pm$0.71 & 65.61$\pm$0.96 & 72.47$\pm$0.84 & 49.16$\pm$0.72 &	56.31$\pm$0.71\\
			
			\bottomrule
	\end{tabular}}
\end{table*}

\begin{table*}
	\footnotesize
	\caption{Average classification accuracies on cifar-100 with different approaches in single-source form.}
	\label{tab:nlu:single2}
	\centering
	\makebox[\textwidth]{
		\begin{tabular}{cccc|cccc}
			\toprule
			\multirow{2}{*}{\textbf{Model}} & \multirow{2}{*}{\textbf{Model Type}} & \multicolumn{2}{c|}{\textbf{5 class}} & \multicolumn{2}{c}{\textbf{5 vs. 3}} & \multicolumn{2}{c}{\textbf{3 vs. 5}}\\
			\cline{3-8}
			& & 2-shot & 4-shot & 2-shot & 4-shot & 2-shot & 4-shot\\
			\midrule
			Meta-SGD & - & 58.88$\pm$0.92 & 65.47$\pm$0.83 & - & - & - & - \\
			\midrule
			Matching Network & Basic (No FCE) & 61.83$\pm$0.89 & 67.43$\pm$0.83 & 73.88$\pm$1.13 & 77.69$\pm$0.99 & 57.61$\pm$0.89 & 64.06$\pm$0.84\\
			Matching Network fine-tune & Basic (No FCE) & 62.38$\pm$0.91 & 67.78$\pm$0.79 & \textbf{75.97$\pm$1.05} & \textbf{79.63$\pm$0.93} & \textbf{59.59$\pm$0.89} & \textbf{66.42$\pm$0.81}\\
			\textbf{Meta Matching Network} & Basic (No FCE) & \textbf{63.07$\pm$0.87} & \textbf{68.55$\pm$0.83} & 74.77$\pm$1.11 & 79.37$\pm$0.95 & 59.09$\pm$0.93 & 65.91$\pm$0.82\\
			
			\midrule
			Prototypical Network & Euclid. & 56.89$\pm$0.86	 & 65.17$\pm$0.85 & 70.65$\pm$1.14 & 75.13$\pm$1.02 & 55.48$\pm$0.95 & 62.94$\pm$0.85\\
			Prototypical Network fine-tune & Euclid. & \textbf{59.25$\pm$0.88} & 65.01$\pm$0.81 & \textbf{72.85$\pm$1.11} & \textbf{77.17$\pm$0.91} & \textbf{57.70+-0.88} & \textbf{64.22+-0.82}\\
			\textbf{Meta Prototypical Network} & Euclid. & 59.21$\pm$0.91 & \textbf{66.23$\pm$0.81} & 70.94$\pm$1.14 & 76.29$\pm$0.96 & 55.11$\pm$0.93 &	62.74$\pm$0.83\\
			
			\bottomrule
	\end{tabular}}
\end{table*}

\begin{table*}
	\footnotesize
	\caption{Average classification accuracies on caltech-256 with different approaches in single-source form.}
	\label{tab:nlu:single3}
	\centering
	\makebox[\textwidth]{
		\begin{tabular}{cccc|cccc}
			\toprule
			\multirow{2}{*}{\textbf{Model}} & \multirow{2}{*}{\textbf{Model Type}} & \multicolumn{2}{c|}{\textbf{5 class}} & \multicolumn{2}{c}{\textbf{5 vs. 3}} & \multicolumn{2}{c}{\textbf{3 vs. 5}}\\
			\cline{3-8}
			& & 2-shot & 4-shot & 2-shot & 4-shot & 2-shot & 4-shot\\
			\midrule
			Meta-SGD & - & 58.60$\pm$0.81 & 66.49$\pm$0.72 & - & - & - & - \\
			\midrule
			Matching Network & Basic (No FCE) & 62.40$\pm$0.83 & 68.63$\pm$0.72 & 73.01$\pm$0.93 & 78.93$\pm$0.83 & 59.49$\pm$0.82 & 66.99$\pm$0.73\\
			Matching Network fine-tune & Basic (No FCE) & \textbf{63.73$\pm$0.82} & 69.42$\pm$0.70 & \textbf{75.50$\pm$0.90} & \textbf{80.65$\pm$0.82} & 61.05$\pm$0.83 & 67.56$\pm$0.71\\
			\textbf{Meta Matching Network} & Basic (No FCE) & 63.32$\pm$0.84 & \textbf{70.13$\pm$0.72} & 74.14$\pm$0.95 & 80.13$\pm$0.82 & \textbf{61.06$\pm$0.80} & \textbf{67.65$\pm$0.71}\\
			
			\midrule
			Prototypical Network & Euclid. & 58.93$\pm$0.83	 & 67.90$\pm$0.73 & 70.45$\pm$1.02 & 76.99$\pm$0.84 & 56.52$\pm$0.84 & 64.29$\pm$0.77\\
			Prototypical Network fine-tune & Euclid. & 59.62$\pm$0.80 & 68.92$\pm$0.71 & \textbf{73.14$\pm$0.99} & \textbf{79.42$\pm$0.82} & \textbf{59.78$\pm$0.82} & 65.32$\pm$0.75\\
			\textbf{Meta Prototypical Network} & Euclid. & \textbf{60.28$\pm$0.81} & \textbf{68.99$\pm$0.71} & 71.10$\pm$1.00 & 78.77$\pm$0.84 & 57.82$\pm$0.84 &	\textbf{65.99$\pm$0.76}\\
			
			\bottomrule
	\end{tabular}}
\end{table*}

\begin{table*}
	\footnotesize
	\caption{Average classification accuracies on cub-200 with different approaches in single-source form.}
	\label{tab:nlu:single4}
	\centering
	\makebox[\textwidth]{
		\begin{tabular}{cccc|cccc}
			\toprule
			\multirow{2}{*}{\textbf{Model}} & \multirow{2}{*}{\textbf{Model Type}} & \multicolumn{2}{c|}{\textbf{5 class}} & \multicolumn{2}{c}{\textbf{5 vs. 3}} & \multicolumn{2}{c}{\textbf{3 vs. 5}}\\
			\cline{3-8}
			& & 2-shot & 4-shot & 2-shot & 4-shot & 2-shot & 4-shot\\
			\midrule
			Meta-SGD & - & 57.18$\pm$0.81 & 62.55$\pm$0.76 & - & - & - & - \\
			\midrule
			Matching Network & Basic (No FCE) & 56.92$\pm$0.81 & 61.99$\pm$0.77 & 70.90$\pm$1.06 & 75.43$\pm$0.98 & 56.18$\pm$0.82 & 61.55$\pm$0.77\\
			Matching Network fine-tune & Basic (No FCE) & \textbf{58.73$\pm$0.78} & \textbf{64.14$\pm$0.72} & \textbf{71.69$\pm$0.98} & \textbf{76.37$\pm$0.86} & \textbf{57.70$\pm$0.81} & \textbf{62.45$\pm$0.73}\\
			\textbf{Meta Matching Network} & Basic (No FCE) & 58.14$\pm$0.82 & 62.98$\pm$0.74 & 70.95$\pm$1.02 & 75.38$\pm$0.92 & 56.50$\pm$0.79 & 61.63$\pm$0.74\\
			
			\midrule
			Prototypical Network & Euclid. & 55.77$\pm$0.86	 & 63.13$\pm$0.75 & 69.66$\pm$1.05 & 76.05$\pm$0.96 & 54.20$\pm$0.83 & 60.30$\pm$0.74\\
			Prototypical Network fine-tune & Euclid. & \textbf{57.21$\pm$0.81} & \textbf{64.01$\pm$0.77} & \textbf{72.47$\pm$0.99} & 76.74$\pm$0.89 & 54.34$\pm$0.80 & \textbf{62.99$\pm$0.73}\\
			\textbf{Meta Prototypical Network} & Euclid. & 55.58$\pm$0.83 & 63.49$\pm$0.73 & 68.73$\pm$1.05 & \textbf{77.04$\pm$0.90} & \textbf{54.50$\pm$0.81} &	60.91$\pm$0.72\\
			
			\bottomrule
	\end{tabular}}
\end{table*}

\paragraph{Baseline Models} There are three baseline models in our experiments:  Matching Network, Prototypical Network, and Meta-SGD with CNN as the base model. For Matching Network and Prototypical Network, we implement our own versions. We only implement Matching Network without fully-conditional embedding (FCE). We choose Euclidean distance in Prototypical Network as is suggested in \cite{NIPS2017_6996}. For Meta-SGD, we extend the version of \cite{MAMLFinnAL17} to support all of the four datasets. 

\paragraph{CNN architectures} The CNN architecture in \cite{vinyals2016matching} and \cite{NIPS2017_6996,s3pool} is used, which consists of 4 modules with a $3 \times 3$ convolution with 64 filters followed by batch normalization, a ReLu non-linearity and $2 \times 2$ max-pooling. In \cite{vinyals2016matching} and \cite{NIPS2017_6996}, dropout is not used. Here we use dropout with a small rate 0.1 in our Meta-Metric-Learner to reduce over-fitting in our experiment. For all models, the loss function is the classification cross-entropy between the predicted and true class.

\paragraph{Hyper-parameters} There are several hyper-parameters required for our Meta-Metric-Learner and baseline models, including dropout rate, learning rate of the meta-learner, and the number of gradient steps $N$. All of them are tuned in the meta-validation set. 

\subsection{Experiments in Single-Source Form}
To demonstrate the effectiveness of our Meta-Metric-Learner, we first execute experiments for single task/resource on all of the four image datasets, in which no auxiliary set $\mathscr{D}_{aux}$ is available from other tasks. Thus we need to perform $k=m*2$-shot learning ($m=1,2$), i.e., for each class, we split its samples into two parts equally, which are used as query samples and supporting samples respectively to calculate the gradient of the base learner. We test our approach in the classic ``$k$-shot $N$-way''  few-shot learning setting following previous works
and a new but more realistic few-shot setting with flexible class numbers. We randomly construct 800 few-shot learning tasks as $\mathscr{D}_{meta-train}$ for meta-training, 600 tasks as $\mathscr{D}_{meta-validate}$ for validation and 600 tasks as $\mathscr{D}_{meta-test}$ for performance evaluation. For the classic ''$k$-shot $N$-way'' few-shot learning setting, each task contains images of 5 different classes, each with 2 or 4 samples in the supporting set and 15 samples in the testing set. Because numbers of classes in $\mathscr{D}_{meta-train}$ and $\mathscr{D}_{meta-test}$ are the same, the original Meta-SGD method can be employed. So in this setting, our baseline models are Meta-SGD, Matching Network, and Prototypical Network. For the second setting, we test our model in two modes: 1) 5 and 3 classes for meta-training and meta-testing,  2) 3 and 5 classes for meta-training and meta-testing, similarly with 2 or 4 samples in the supporting set and 15 samples in the testing set from each class. Note that 2) is a more challenging setting since the number of classes in $\mathscr{D}_{meta-train}$ is smaller than that in $\mathscr{D}_{meta-test}$. The Meta-SGD method can't be implemented directly in this setting, so our baseline models are only Matching Network and Prototypical Network. For all of the settings, tasks in the meta-validate dataset have the same number of classes as those in the meta-test dataset. We use meta-validate set to adjust the hyper-parameters. To have a fair comparison, all the baselines trained with 2 or 4 samples per class according to their own recipes. We test two kinds of Meta-Metric-Learner, i.e. Meta Matching Network (MMN) model and Meta Prototypical Network (MPN) model. Since the idea of \cite{NIPS2018Tyler} is quite similar to ours, we also implement and test thier model in our own setting, named Matching Network Finetune and Prototypical Network fine-tune in our experiments. To fine-tune the metric model, we split the supporting set equally, the same way as in our MMN and MPN model.

In our experiments, we find that increasing the number of gradient steps within a certain range can improve the model performance. But it won't help a lot if the number of gradient steps is set to large, and a large number of gradient steps will increase the computational complexity of our model. We set gradient steps of training and testing to 5 for the MMN model and 7 for the MPN model. Both models are trained with task batch size of 4 and the learning rate of the meta-learner parameters is set to 0.001. All models are trained for 30000 iterations on a single NVIDIA GeForce GTX 1080ti GPU. We make an evaluation every 500 training iterations and record the best testing accuracy during the training procedure as the final result.

The results are shown in Table \ref{tab:nlu:single1} to \ref{tab:nlu:single4}. 
All the results are averaged over 600 tasks from $\mathscr{D}_{meta-test}$ with $95\%$ confidence intervals. From these tables, we can see that Matching Network performs better than Prototypical Network. Perhaps this is because the number of classes in $\mathscr{D}_{meta-train}$ and $\mathscr{D}_{meta-test}$ are set fixed beforehand, and not tuned on a held-out validation set like \cite{NIPS2017_6996}. And in the case where there is a finite number of meta-train tasks, the original Meta-SGD method seems to perform worse than the plain Matching Network. Moreover, it is obvious to see that the 3 vs. 5 split is a more challenging task. Comparing the results in both cases, the performance of 3 vs. 5 is around 15\% lower than 5 vs. 3 cases. In both settings, we can see that task-specific metric models, no matter finetuned or meta-learned, outperform the baseline models. When comparing our method to \cite{NIPS2018Tyler}, different models perform better in  different scenarios, showing that our method is more effective to some degree. 

\begin{figure*}
\centering
\subfigure[Training procedure of models related to Matching Network]{\includegraphics[scale=0.39]{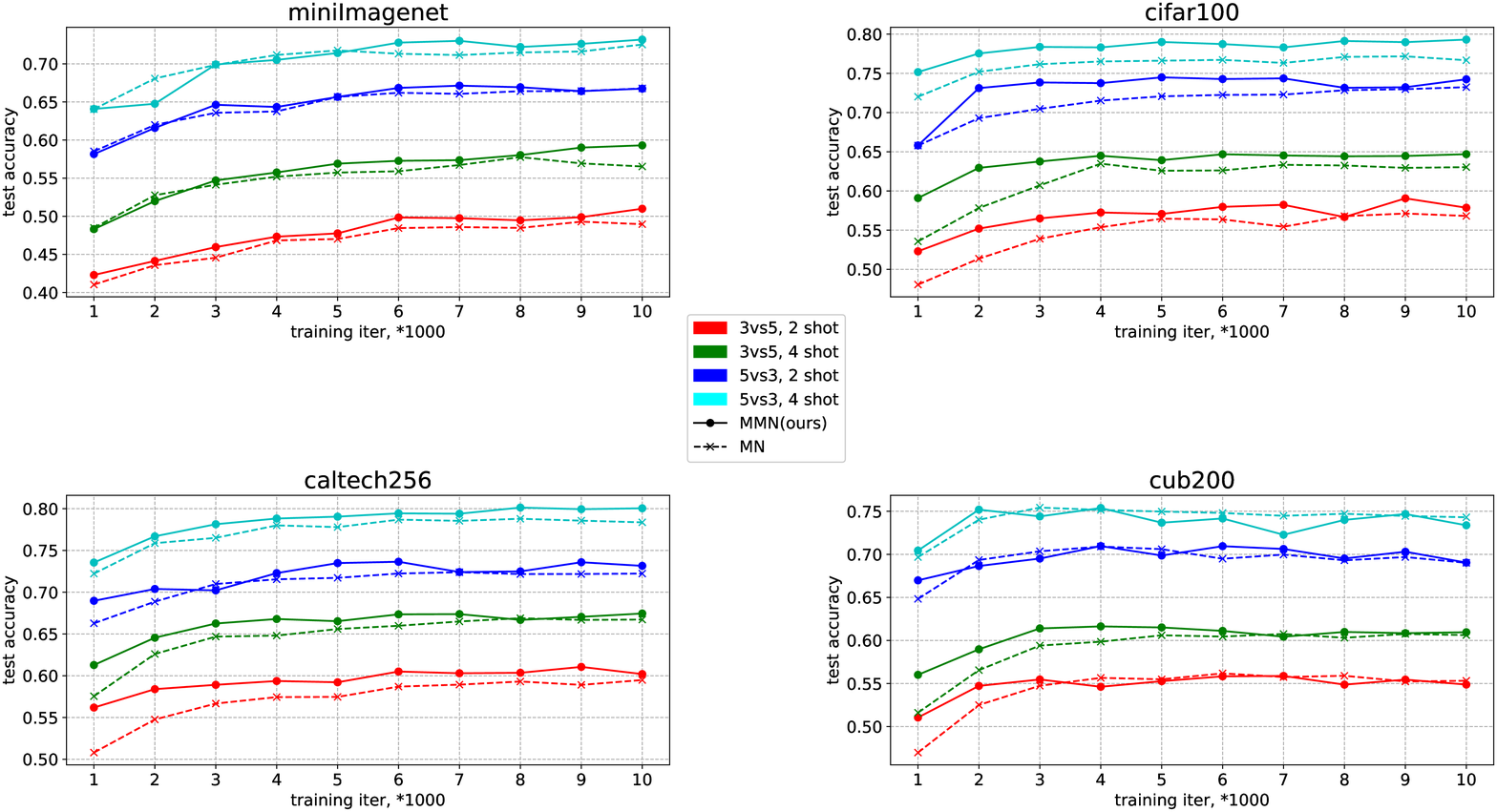}
\label{fig:result:a}
}

\subfigure[Training procedure of models related to Prototypical Network]{\includegraphics[scale=0.39]{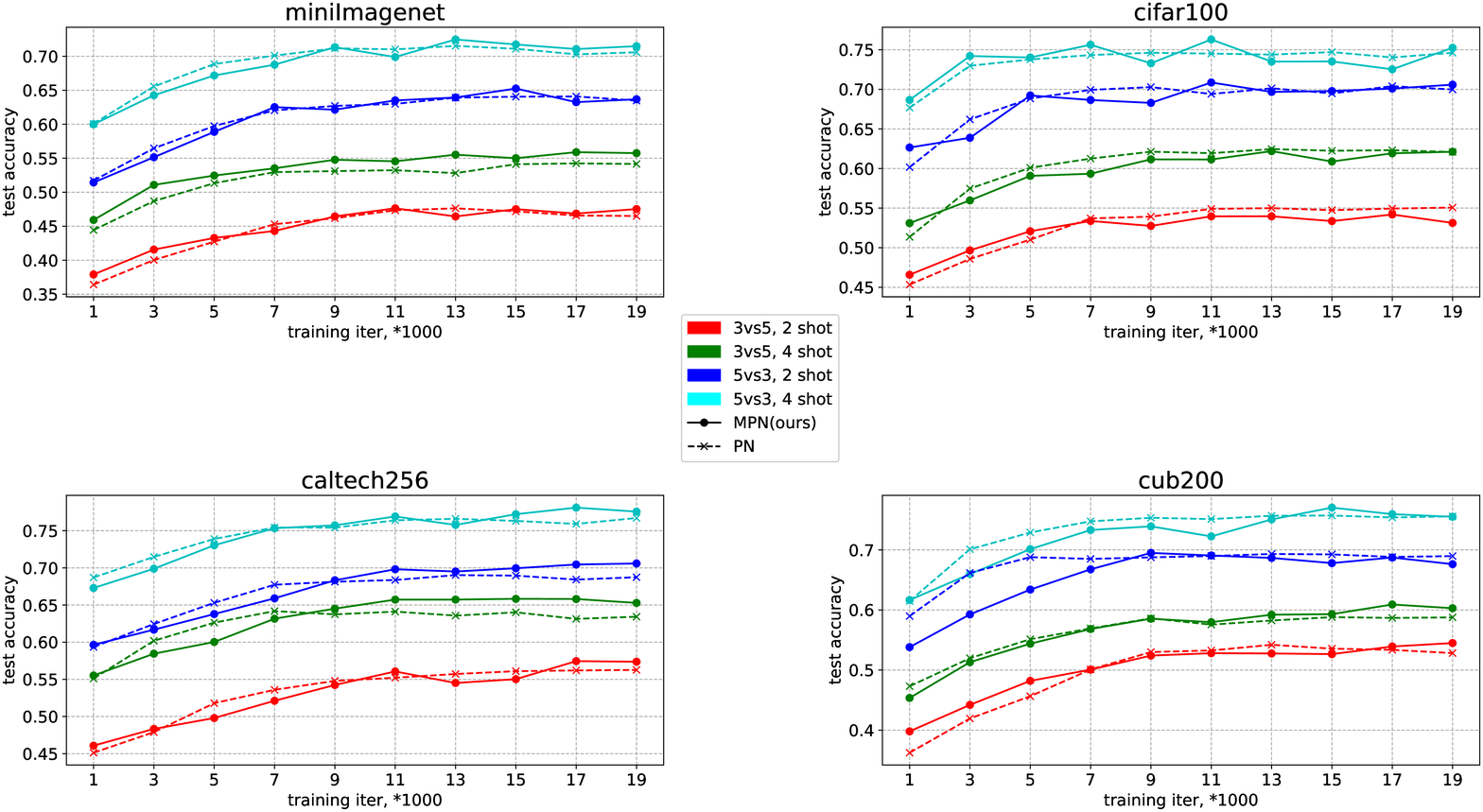}
\label{fig:result:b}
}

\caption{The relationship between test accuracy and training iteration steps in the case of unbalanced numbers of classes. Different colors represent different experiment settings. Solid lines with dot markers show the results of our models and dash lines with cross markers show the results of corresponding baseline models.}
\label{fig:result:single}
\end{figure*}

Figure \ref{fig:result:single} shows the relationship between test accuracy and training iteration steps in the setting of unbalanced numbers of classes. We can see that our methods converge faster and better than corresponding baseline models in most of the cases. For some other cases, our methods can still achieve better test accuracy although there exists some fluctuation compared with baseline models during training.

\subsection{Experiments in Multi-Source Form}
In this section, we show the results of our Meta-Metric-Learner when there is an auxiliary set $\mathscr{D}_{aux}$ available. We test this setting on Omniglot dataset because it naturally forms a multi-source setting. Here each data source corresponds to an alphabet; and the motivation of multi-source setting is to explore the cross-alphabet knowledge sharing to boost the performance on a target alphabet.

\begin{table*}
	\footnotesize
	\caption{Average classification accuracies on Omniglot with different approaches in multi-source form.}
	\label{tab:omn:result}
	\centering
	\makebox[\textwidth]{
		\begin{tabular}{cccc}
			\toprule
			\textbf{Model} & \textbf{Model Type} & \textbf{30\% classes, 1-shot} & \textbf{30\% vs  50\%, 1-shot} \\
			
			\midrule
			Meta-SGD & - & 81.75\% & - \\
			
			\midrule
			Matching Network & Basic (No FCE) & 84.92\% & 76.69\%     \\ 
			\textbf{Meta Matching Network} & Basic (No FCE) & \textbf{85.43\%} & \textbf{77.30\%}  \\
			
			\midrule
			Prototypical Network & Euclid. & 86.58\% & 78.83\% \\ 
			\textbf{Meta Prototypical Network} & Euclid. & \textbf{86.84\%} & \textbf{79.25\%}  \\
			\bottomrule
		\end{tabular}
	}
\end{table*}

In the experiment, the total number of data sources (i.e. alphabets) is 50.  We randomly choose 10 alphabets as target data source and 30 alphabets as the auxiliary data source in this experiment. We use the strategy introduced above to find related top-$s$ ($s$ is different according to different sources and 1 or 2 just works fine in our experiment) sources and train the Meta-Metric-Learner. We still test our model in both identical class number setting and flexible class number setting. In the first setting, every meta-dataset contains 30\% of classes in the data source. In the second setting, we split classes in the data source with 3:2:5 as meta-train, meta-validate, and meta-test. We only conduct experiments in 1-shot learning mode. There are 10 examples per class for evaluation in the testing set of each task. We follow the procedure of \cite{vinyals2016matching} by augmenting the characters with rotations in multiples of 90 degrees. Average classification accuracy of the 10 target data source are shown in Table \ref{tab:omn:result}. For all of the settings, our model can achieve better classification accuracy than others.

\section{Conclusion}
In this paper, we propose the Meta-Metric-Learner for few-shot learning, which is a combination of a Meta-SGD meta-learner and a base metric classifier. The proposed method takes several advantages such as being able to handle unbalanced classes as well as to generate task-specific metrics. Moreover, using the meta-learner to guide gradient optimization in metric learners seems to be a promising direction. We evaluate our model on several datasets, in both single-source and multi-source settings. The experimental results demonstrate that our approach is effective for few-shot learning problems. 

There are several directions for future work. First, we will exploit more sophisticated optimization-based meta-learning models such as probabilistic MAML and LEO to adjust the basic metric-based model so that we can make the learned metric generalize better. Secondly, we would like to focus on selecting data from more related domains/sources to support the training of Meta-Metric-Learners. Thirdly, it would be interesting to propose an end-to-end framework of the Meta-Metric-Learner to leverage the data from different domains/sources/tasks for the training, instead of the current two-stage procedure. Besides, we would like to move forward to apply the current framework in other applications, such as language modeling and attribute classification \cite{LuKZCJF17}. 

\ifCLASSOPTIONcaptionsoff
  \newpage
\fi

\bibliographystyle{IEEEtran}
\bibliography{main}

\end{document}